\documentclass[10pt, conference, compsocconf]{IEEEtran}

\IEEEoverridecommandlockouts

\usepackage{mathptmx} 
\usepackage{cite}
\usepackage{amsmath,amssymb,amsfonts}
\usepackage{graphicx}
\usepackage{textcomp}
\usepackage{xcolor}
\usepackage[numbers]{natbib}

\usepackage{fancyhdr}
\usepackage[normalem]{ulem}
\usepackage{listings}
\usepackage{float}
\usepackage{tabularx}
\usepackage{xargs}                      

\usepackage{caption}
\usepackage{float}
\usepackage{placeins}
  
\newcolumntype{L}{>{\RaggedRight\arraybackslash}X}
\newcolumntype{C}{>{\centering\arraybackslash}X}

\newcommand{\ignore}[1]{}

\usepackage[disable]{todonotes} 

\newif\ifsubmit
\submittrue
\ifsubmit
    \newcommand{\usure}[1]{}
    \newcommand{\change}[1]{}
    \newcommand{\info}[1]{}
    \newcommand{\improvement}[1]{}
    \newcommandx{\cheng}[2][1=]{}
    \newcommandx{\abdul}[2][1=]{}
    \newcommandx{\simon}[2][1=]{}
    \newcommandx{\junjun}[2][1=]{}
    \newcommandx{\wenmei}[2][1=]{}
\else
	 
	\newcounter{adtodocounter} 
	\newcounter{cltodocounter} 
	\newcounter{jjtodocounter} 
	\newcounter{wmhtodocounter} 
	\newcounter{sgtodocounter} 
    \newcommandx{\unsure}[2][1=]{\todo[linecolor=red,backgroundcolor=red!25,bordercolor=red,#1]{#2}}
    \newcommandx{\change}[2][1=]{\todo[linecolor=blue,backgroundcolor=blue!25,bordercolor=blue,#1]{#2}}
    \newcommandx{\info}[2][1=]{\todo[linecolor=OliveGreen,backgroundcolor=OliveGreen!25,bordercolor=OliveGreen,#1]{#2}}
    \newcommandx{\improvement}[2][1=]{ \marginpar[\todo[linecolor=Plum,backgroundcolor=Plum!25,bordercolor=Plum,#1]{#2}]{}}
    \newcommandx{\thiswillnotshow}[2][1=]{\todo[disable,#1]{#2}}
    \newcommandx{\cheng}[2][1=]{\stepcounter{cltodocounter} \todo[linecolor=red,backgroundcolor=purple!25,bordercolor=purple,#1]{CL(\thecltodocounter): #2}}
    \newcommandx{\abdul}[2][1=]{\stepcounter{adtodocounter} \todo[linecolor=red,backgroundcolor=red!25,bordercolor=red,#1]{AD(\theadtodocounter): #2}}
    \newcommandx{\simon}[2][1=]{\stepcounter{sgtodocounter} \todo[linecolor=OliveGreen,backgroundcolor=OliveGreen!25,bordercolor=OliveGreen,#1]{SG(\thesgtodocounter): #2}}
    \newcommandx{\jinjun}[2][1=]{\stepcounter{jjtodocounter} \todo[linecolor=yellow,backgroundcolor=yellow!25,bordercolor=yellow,#1]{JJ(\thejjtodocounter): #2}}
    \newcommandx{\wenmei}[2][1=]{\stepcounter{wmhtodocounter} \todo[linecolor=Plum,backgroundcolor=Plum!25,bordercolor=Plum,#1]{WMH(\thewmhtodocounter): #2}}
\fi

\usepackage{blindtext}
\usepackage{amsmath}
\usepackage{array}
\usepackage{pgfplots}
\usepackage{pgfplotstable}
\usepackage{xspace}
\usepackage{multirow}
\usepackage{comment}
\usepackage{array}
\usepackage{makecell}
\usepackage{rotating}
\usepackage{filecontents}
\usepackage{tikz}

\usepackage{amssymb}
\usepackage{pifont}
\usepackage{tcolorbox}
\usepackage{environ}

\newcommand{\cmmnt}[1]{\ignorespaces}

\definecolor{myred}{rgb}{0.843137,0.188235,0.152941}
\definecolor{myblack}{rgb}{0.27451,0.32549,0.384314}
\definecolor{mygreen}{rgb}{0.301961,0.686275,0.290196}
\definecolor{myyellow}{rgb}{0.996078,0.878431,0.564706}
\definecolor{myblue}{rgb}{0.568627,0.74902,0.858824}

\pgfplotsset{compat=newest,}
\usepgfplotslibrary{external}
\usetikzlibrary{babel}

\definecolor{dkgreen}{rgb}{0,0.6,0}
\definecolor{gray}{rgb}{0.5,0.5,0.5}
\definecolor{mauve}{rgb}{0.58,0,0.82}

\usepackage{fancyhdr}
\usepackage{hyperref}

\fancypagestyle{firstpage}{
  \fancyhf{}
\setlength{\headheight}{10pt}
  \pagenumbering{arabic}
}

\tcbuselibrary{listings,breakable}

\pgfplotsset{every axis/.style={scale only axis}}

\definecolor{plotcolor1}{rgb}{0.568627,0.74902,0.858824}
\definecolor{plotcolor2}{rgb}{0.996078,0.878431,0.564706}
\definecolor{plotcolor3}{rgb}{0.27451,0.32549,0.384314}
\definecolor{plotcolor4}{rgb}{0.843137,0.188235,0.152941}
\definecolor{plotcolor5}{rgb}{0.988235,0.552941,0.34902}
\definecolor{plotcolor6}{rgb}{0.596078,0.305882,0.639216}
\definecolor{plotcolor7}{rgb}{0.65098,0.337255,0.156863}
\definecolor{plotcolor8}{rgb}{0.105882,0.619608,0.466667}
\definecolor{plotcolor9}{rgb}{1.,1.,0.6}
\definecolor{plotcolor10}{rgb}{0.745098,0.729412,0.854902}
\definecolor{plotcolor11}{rgb}{0.984314,0.501961,0.447059}
\definecolor{plotcolor12}{rgb}{0.501961,0.694118,0.827451}
\definecolor{plotcolor13}{rgb}{1.,1.,0.2}
\definecolor{plotcolor14}{rgb}{0.992157,0.705882,0.384314}
\definecolor{plotcolor15}{rgb}{0.988235,0.803922,0.898039}
\definecolor{plotcolor16}{rgb}{0.701961,0.870588,0.411765}
\definecolor{plotcolor17}{rgb}{0.215686,0.494118,0.721569}
\definecolor{plotcolor18}{rgb}{0.941176,0.231373,0.12549}
\definecolor{plotcolor19}{rgb}{0.168627,0.54902,0.745098}
\definecolor{plotcolor20}{rgb}{0.552941,0.827451,0.780392}
\definecolor{plotcolor21}{rgb}{0.968627,0.505882,0.74902}
\definecolor{plotcolor22}{rgb}{0.6,0.6,0.6}
\definecolor{plotcolor23}{rgb}{0.301961,0.686275,0.290196}
\definecolor{plotcolor24}{rgb}{0.980392,0.501961,0.447059}

\usepgfplotslibrary{colorbrewer}
\pgfplotsset{cycle list/Dark2-8}

\DeclareRobustCommand{\importantbox}[2][gray!20]{%
\begin{tcolorbox}[   
        breakable,
        left=0pt,
        right=0pt,
        top=0pt,
        bottom=0pt,
        colback=#1,
        colframe=#1,
        width=\dimexpr\columnwidth\relax, 
        enlarge left by=0mm,
        boxsep=5pt,
        arc=0pt,outer arc=0pt,
        ]
        #2
\end{tcolorbox}
}

\newcommand{\floor}[1]{\left\lfloor #1 \right\rfloor}

\newcommand{\pitfall}[1]{\importantbox[myred!50]{#1}}

\newcommand{\carml}{MLModelScope\xspace}

\usepackage{authblk}

\def\BibTeX{{\rm B\kern-.05em{\sc i\kern-.025em b}\kern-.08em
    T\kern-.1667em\lower.7ex\hbox{E}\kern-.125emX}}
    
\title{Challenges and Pitfalls of Machine Learning Evaluation and Benchmarking}
\date{April 2019}

\begin{document}

\author[1]{\normalsize Cheng Li}
\author[1]{\normalsize Abdul Dakkak}
\author[2]{\normalsize Jinjun Xiong}
\author[3]{\normalsize Wen-mei Hwu}

\affil[ ]{\textit {cli99@illinois.edu,dakkak@illinois.edu,jinjun@us.ibm.com, w-hwu@illinois.edu}}

\affil[1]{%
    \small
  Department of Computer Science \\
  University of Illinois, Urbana-Champaign}
\affil[2]{%
    \small
  IBM Thomas J. Watson Research Center \\
  Yorktown Heights, NY}
\affil[3]{%
    \small
  Department of Electrical and Computer Engineering \\
  University of Illinois, Urbana-Champaign}

\maketitle

\begin{abstract}
An increasingly complex and diverse collection of Machine Learning (ML) models as well as hardware/software stacks, collectively referred to as ``ML artifacts'', are being proposed --- leading to a diverse landscape of ML. These ML innovations proposed have outpaced researchers' ability to analyze, study and adapt them. This is exacerbated by the complicated and sometimes non-reproducible procedures for ML evaluation. A common practice of sharing ML artifacts is through repositories where artifact authors post ad-hoc code and some documentation, but often fail to reveal critical information for others to reproduce their results. This results in users' inability to compare with artifact authors' claims or adapt the model to his/her own use. This paper discusses common challenges and pitfalls of ML evaluation and benchmarking , which can be used as a guideline for ML model authors when sharing ML artifacts, and for system developers when benchmarking or designing ML systems.
 
\end{abstract}

\section{Introduction}

An increasingly complex and diverse collection of ML models as well as hardware/software stacks are being proposed each day.
This has lead to a vibrant and diverse landscape of ML.
The amount of ML solutions are overwhelming. In ~\cite{dean2018new} the authors show that the number of ML \textit{arXiv} papers published has outpaced Moore's law.
Thus researchers struggle to keep up to date and to analyze, study and adapt them. This is exacerbated by the complicated and sometimes non reproducible procedures for ML evaluation.

To facilitate and accelerate the adoption of ML innovations, ML evaluation must be easily reproducible and a better way of sharing ML artifacts is needed.
The current practice of sharing ML artifacts is by publishing source code to repositories such as GitHub. Model authors post their ad-hoc code and some documentation. We often find authors fail to reveal critical information for others to reproduce their results. Some authors also release Dockerfiles. However, Docke only guarantees the software stack but does not help model users examine or modify the artifact to adapt to other environments. In short, one often fails to reproduce artifact authors' claims, not to mention adapt the models to his/her own use. 

This paper discusses the challenges of ML evaluation and benchmarking, and outlines common pitfalls model users often encounter when attempting to replicate model authors' claims. 
In ~\cite{dakkak2018mlmodelscope}, we present \carml, an open-source ML evaluation system which lowers the cost and effort for performing model evaluation and benchmarking, making it easier to reproduce, evaluate, and analyze accuracy, performance, and resilience claims of ML artifacts.
This paper documents some of the lessons we learned when developing \carml, and aims to inform both model authors on the critical information they must reveal for others to reproduce their claims, and system developers on how to benchmark or design a ML system in a reproducible manner. 

\begin{figure}[t]
\centering
\includegraphics[width=0.5\textwidth]{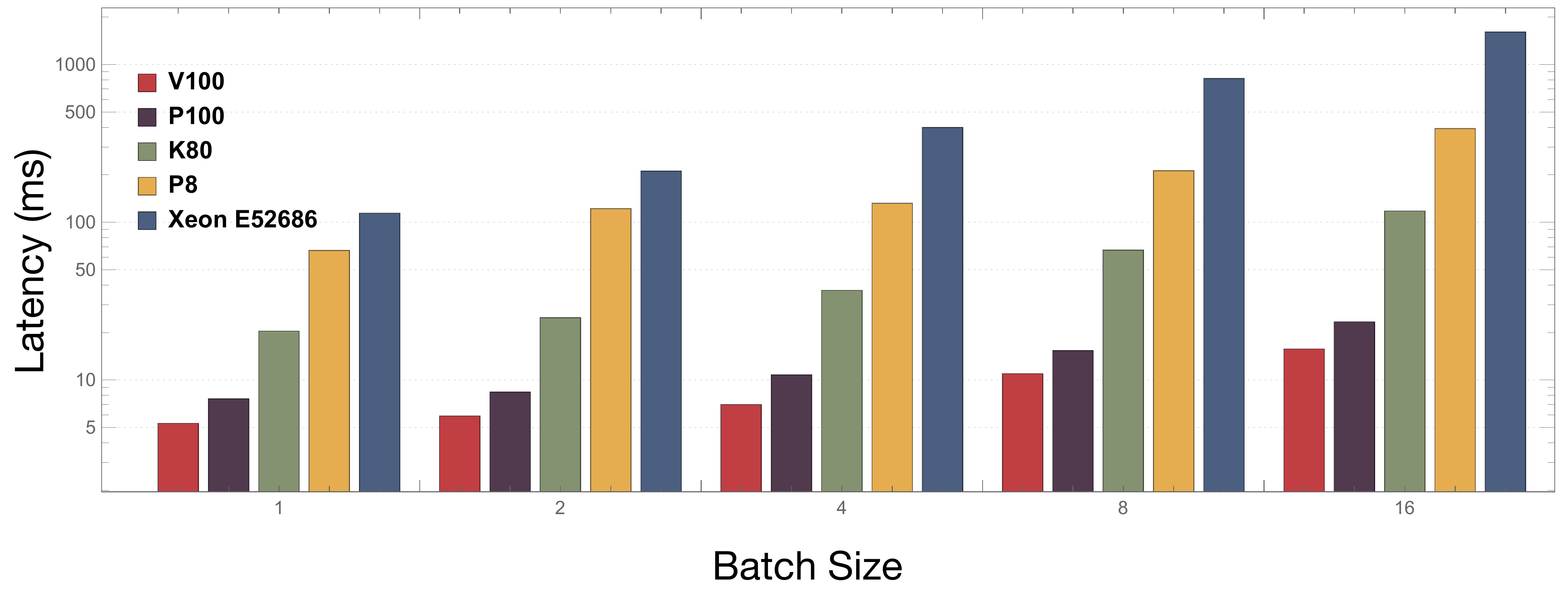}
\caption{ResNet\_v1\_50 using TensorFlow 1.13 on GPU and GPU systems with varying batch sizes.}
\label{fig:cpu_v_gpu}
\end{figure}



\begin{figure*}[t]
\centering
\includegraphics[width=0.9\textwidth]{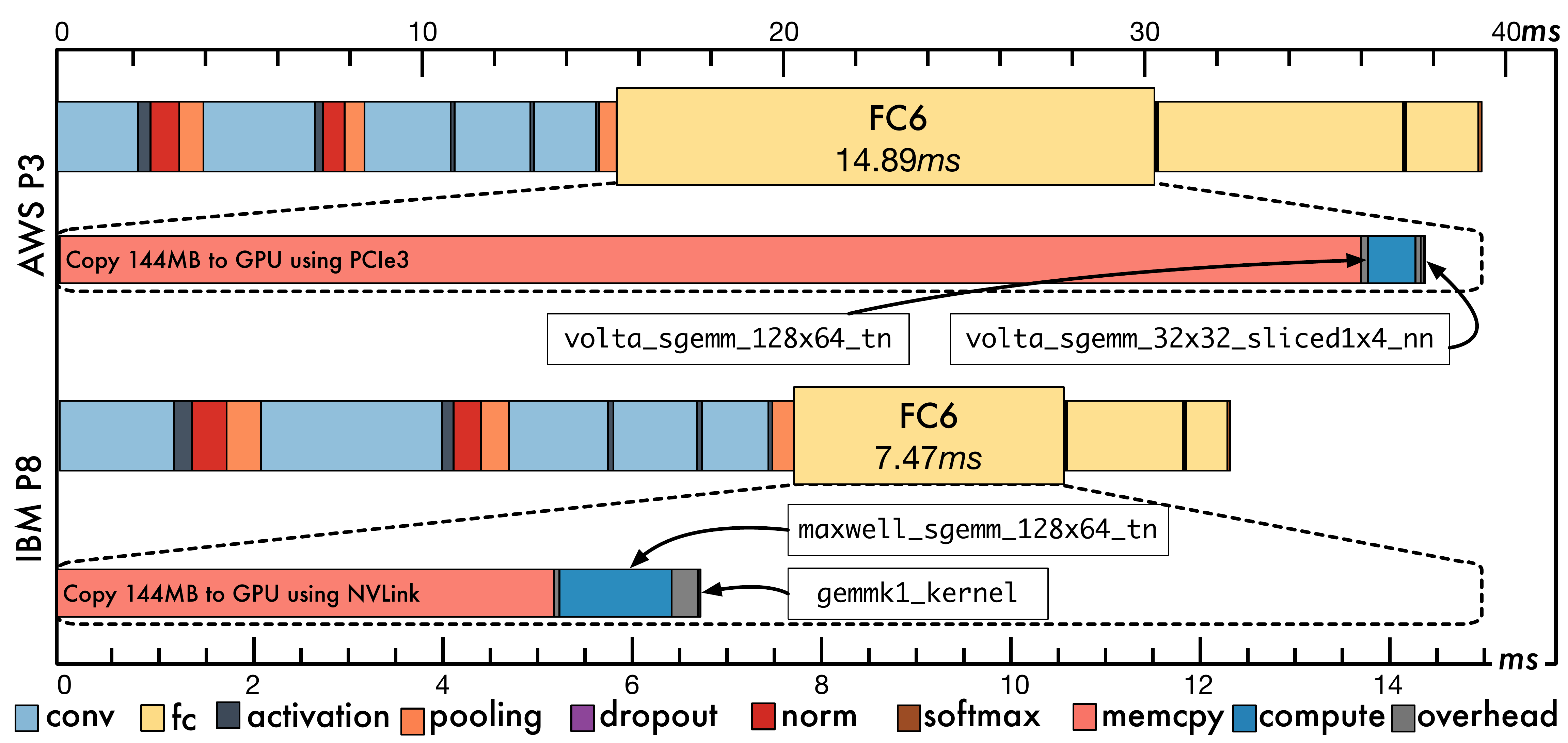}
\caption{POWER8 with Pascal GPU and NVLink vs X86 with Volta for a ``cold-start'' inference using Caffe AlexNet for batch size $64$.
The color coding of layers and runtime functions signify that they have the same kernel implementation, but does not imply that the parameters are the same.}
\label{fig:alexnet_detail}
\end{figure*}

\section{Challenges and Pitfalls}

Complicated and sometimes non-reproducible procedures for ML artifacts is currently a big hurdle in research adoption.
The lack of standard and efficient systems for specifying and provisioning ML evaluation is the main cause of the \textit{pain point}.
There are many factors that must work in unison within a ML model workflow, including hardware, programming language, pre/post-processing, model, dataset, software stack and hardware configurations.
Researchers who publish and share ML artifacts are often unaware of some of the factors, and fail to reveal the information critical for others to reproduce their results.
In the process of developing \carml we identified a few common pitfalls and handled them in the model manifest specification and the platform's design.
This section details the factors that affect ML evaluation,  how the pitfalls arise, and provides suggested solutions.
 
\subsection{Hardware}

Different hardware architectures can result in varying performance and accuracy, since system and ML libraries leverage features within the hardware architecture.

\pitfall{
Pitfall 1: Only look at partial hardware, not the entire system. E.g. Inference on a Volta GPU must be faster than that on a Pascal GPU.}

Figure~\ref{fig:cpu_v_gpu} compares inference performance across systems. Volta (V100) is faster than Pascal (P100) in this case. One often assumes this to be always true. However, looking at only GPU or CPU compute sections when comparing performance is a common pitfall. Figure~\ref{fig:alexnet_detail} shows a Pascal system can perform better than a Volta system because of a faster CPU-GPU interconnect.
One therefore should consider the entire system and its end-to-end latency under different workload scenarios when reporting system performance results.

With \carml's profiling capabilities, one can discern why there is a performance difference. Figure~\ref{fig:alexnet_detail} shows the layer and GPU kernel breakdown of the model inference on the two systems. We ``zoom-into'' the longest running layer (\texttt{FC6}) and show the model inference choke point. The difference between the model performance mainly comes form \texttt{FC6} layer. On identifying this issue, we were able to look at the Caffe source code and observe that Caffe does lazy copy, meaning the layer weights get copied from CPU to GPU only when it's needed. For \texttt{FC6}, $144MB$ of weights needs to be transferred. As we can see in the GPU kernel breakdown, even though the V100 performs better for \texttt{SGEMM} computation, with the NVLink~\cite{nvlink} (faster than PCIe) between CPU and GPU the IBM P8 system achieves higher memory bandwidth and thus achieves a $2\times$ speedup for \texttt{FC6} layer.

\begin{figure}[ht]
\centering
\includegraphics[width=0.45\textwidth]{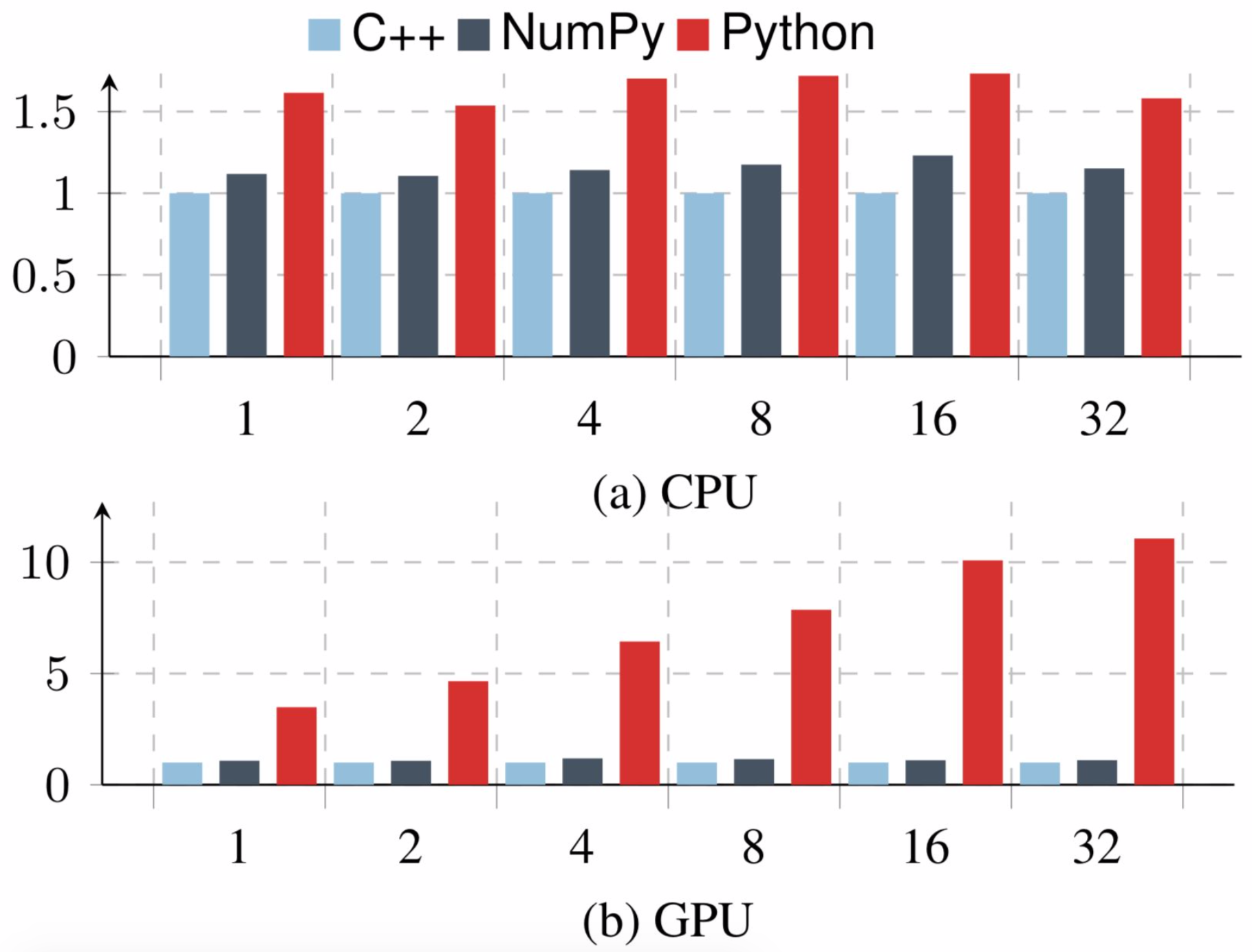}
\caption{Execution time (normalized to C/C++) vs. batch size of Inception-v3 inference on CPU and GPU using TensorFlow with C++, Python using NumPy data types, and Python using native lists.}
\label{fig:c_vs_python}
\end{figure}

\subsection{Programming Language}

Core ML algorithms within frameworks are written in C/C++ for performance and in practice low-latency inference uses C/C+. It is common for developers to use NumPy for numerical computation (NumPy arrays are not Python objects). ML frameworks optimize the execution for NumPy arrays, and avoid memory copy overhead when interfacing with C/C++ code.

\pitfall{Pitfal 2: Use Python API to measure and report bare-metal benchmark results or to deploy latency sensitive production code.}

While no one claims Python to be as fast as C++, we find researchers believe that the glue code that binds Python to C++ takes negligible time. For example, benchmarks such as MLPerf are implemented in Python and report the latency and throughput for Python code. We show in Figure~\ref{fig:c_vs_python} above that the performance difference between Python and C++ in model evaluation is not negligible and one should use C++ for latency sensitive production code or when reporting bare-metal benchmark results.

\begin{figure}
\centering
\includegraphics[width=0.4\textwidth]{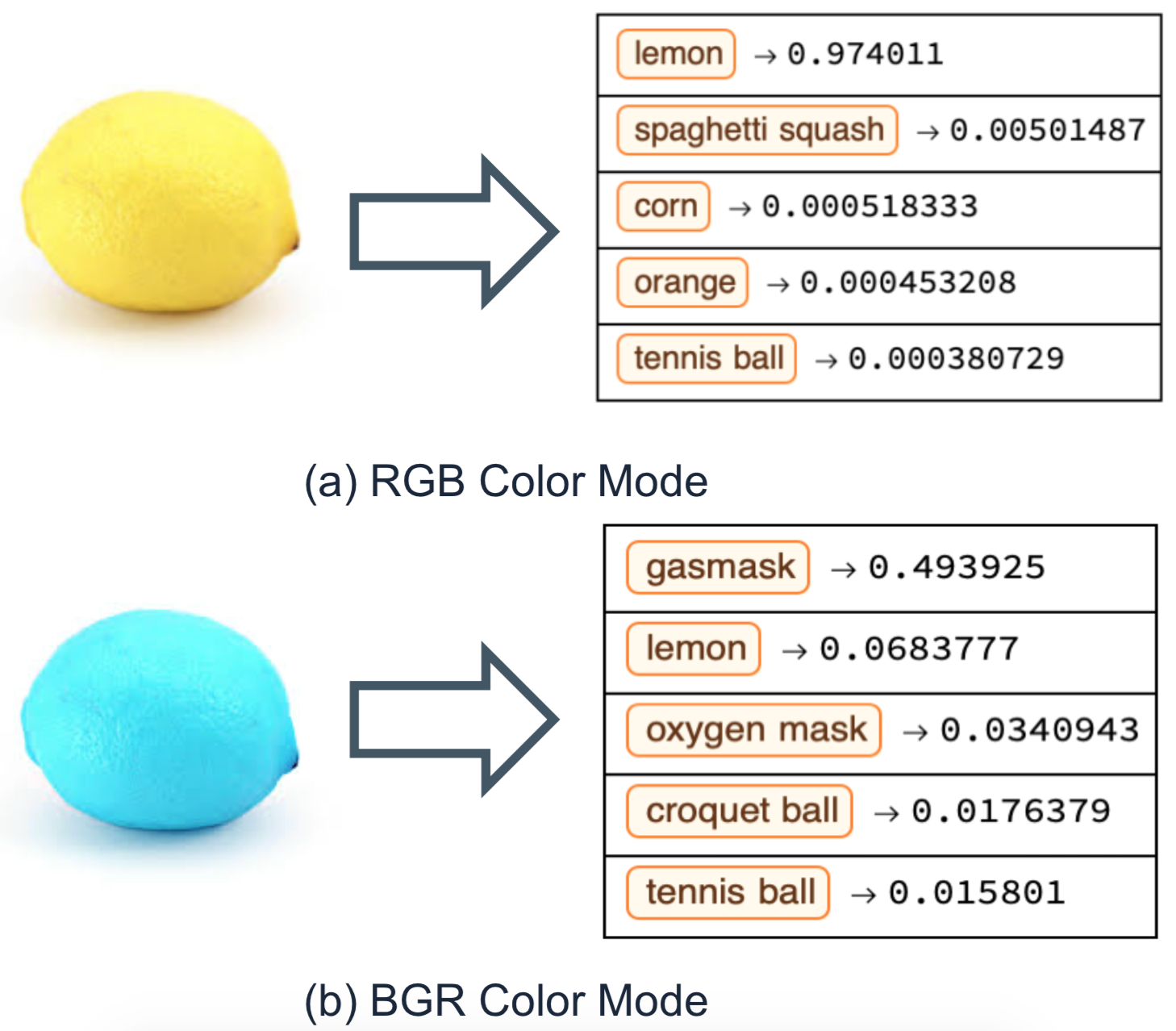}
\caption{Top 5 predictions using Inception-v3 with RGB or BGR color layout.}
\label{fig:lemon}
\end{figure}

\begin{figure}[ht]
\centering
\includegraphics[width=0.4\textwidth]{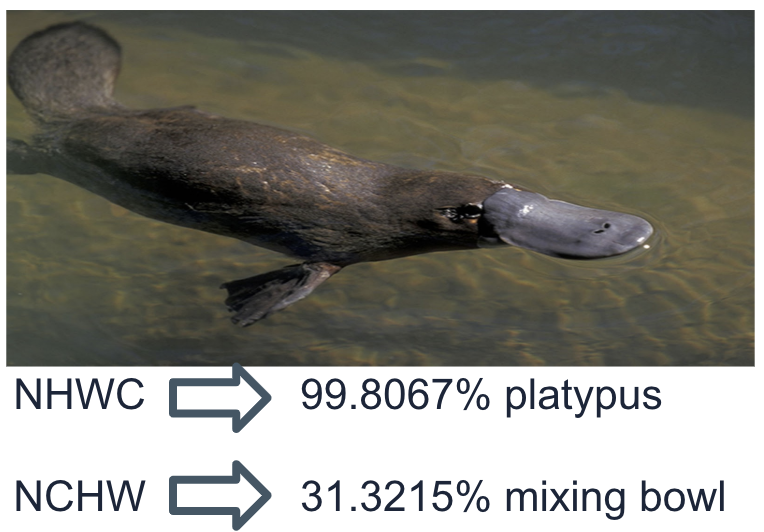}
\caption{Top 1 predictions using Inception-v3 with \texttt{NCHW} or \texttt{NHWC} data layout.}
\label{fig:platapus}
\end{figure}

\subsection{Pre/Post-Processing}

Pre-processing is transforming the user input into a form that can be consumed by the model. Post-processing is processing the model output that can be evaluated using metrics or consumed by subsequent components in the application pipeline. The processing parameters, methods and order affect accuracy and performance.

\pitfall{Pitfall 4: Model authors typically fail to reveal some pre/post-processing details that are needed to reproduce their claims.}

Among all the factors that affect model evaluation accuracy, pre/post-processing is the one that can result in big difference. The input dimension of a model is usually reported by the model author since without the right input dimensions, the model evaluation does not run and gives an error. Even if the input dimension is not explicitly given, model users can inspect the model architecture to figure that out. 

\begin{figure}[ht]
\centering
\includegraphics[width=0.4\textwidth]{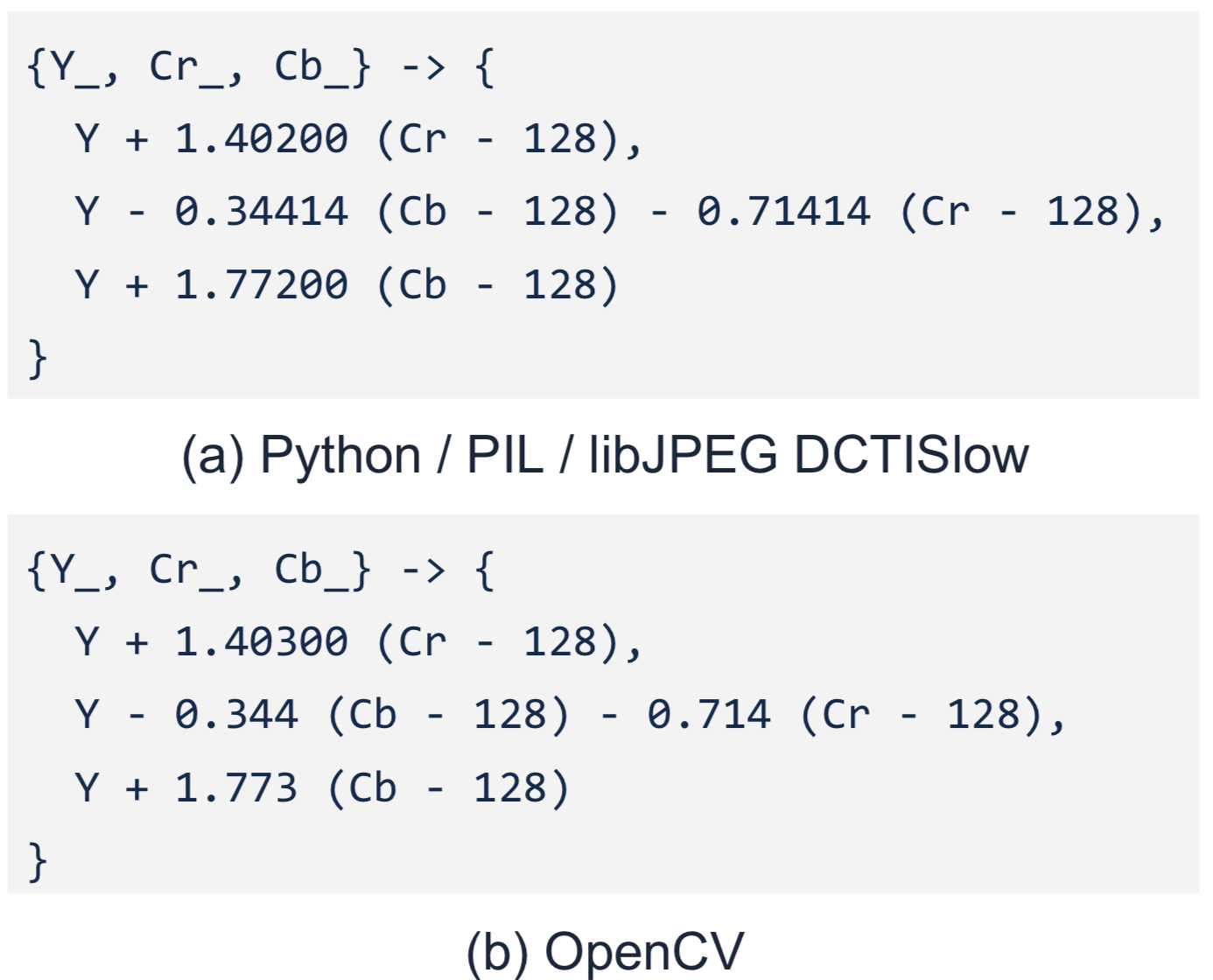}
\caption{Pil vs OpenCV Implementation}
\label{fig:pil_vs_opencv_impl}
\end{figure}

\begin{figure}[ht]
\centering
\includegraphics[width=0.45\textwidth]{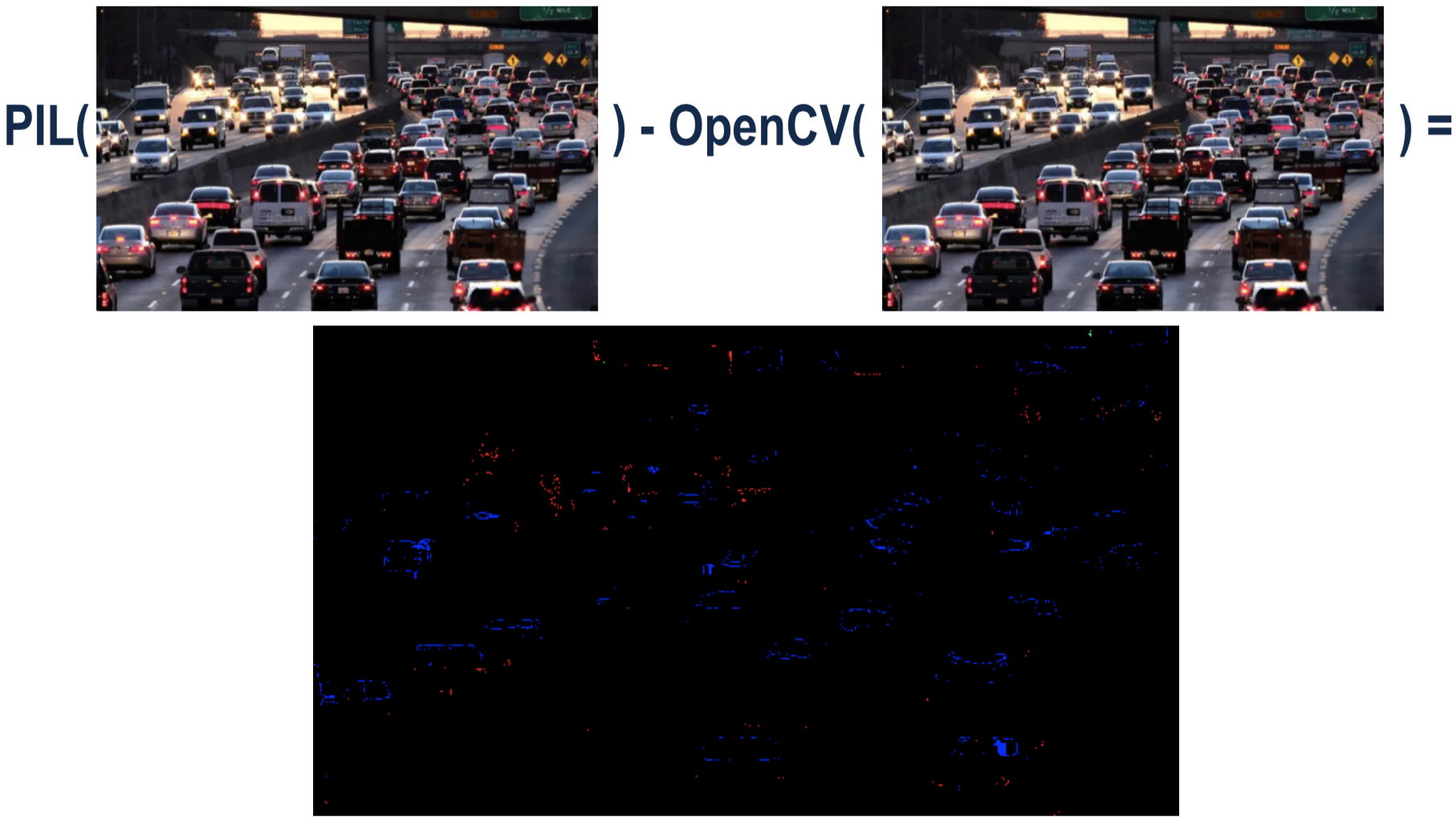}
\caption{Image decoding difference between PIL and OpenCV.}
\label{fig:pil_vs_opencv_diff}
\end{figure}

However, there are some critical pre/post-processing information that if not explicitly reported by the model authors, model users might easily fall into a incorrect evaluation setup and get ``silent errors" in accuracy --- the evaluation runs but the prediction results for some cases are incorrect. These ``silent errors" are difficult to debug. 
Here we take computer vision models as an example and discuss what model users might struggle with when reproducing others' results.

\subsubsection{Color Mode}

Models are trained with decoded images that are in either RGB or BGR color mode. For legacy reasons, OpenCV decodes images in BGR mode by default and subsequently both Caffe and Caffe2 use BGR. Other frameworks such as TensorFlow, PyTorch, MXNet use RGB mode~\cite{caffebgr}.
Figure~\ref{fig:lemon} shows the Inception v3 inference results of the same image using different color modes and everything else being the same.

\begin{figure*}[ht]
\centering
\includegraphics[width=0.9\textwidth]{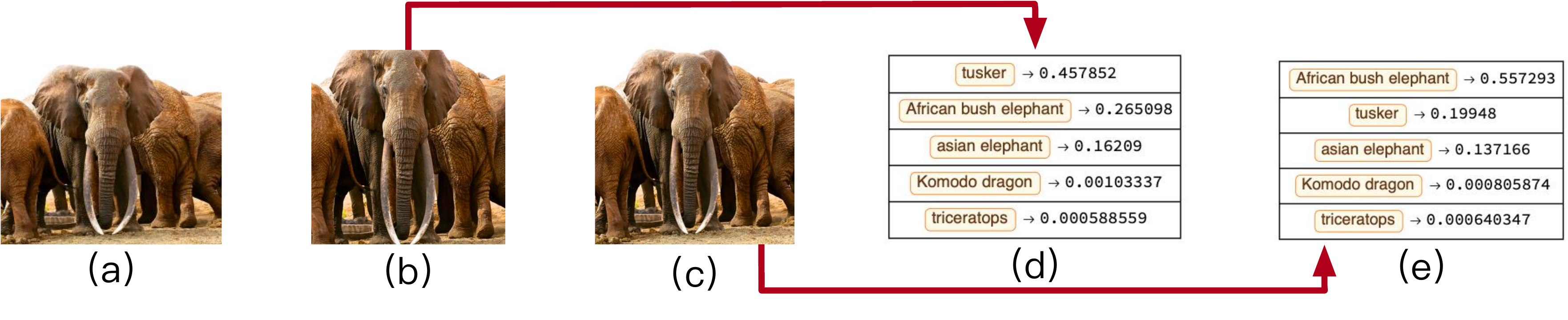}
\vspace{-10pt}
\caption{Differences in the prediction results due to cropping using TensorFlow Inception-v3.}
\label{fig:diff_cropping}
\end{figure*}

\begin{figure*}[ht]
\centering
\includegraphics[width=0.9\textwidth]{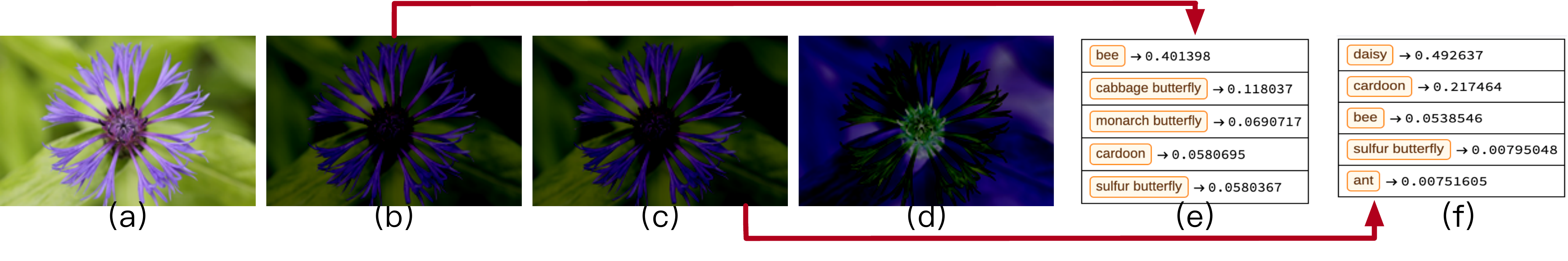}
\caption{Differences due to order of operations for data type conversion and normalization using TensorFlow Inception-v3.}
\label{fig:diff_output_process}
\end{figure*}

\subsubsection{Data Layout}

The data layout for a two-dimensional image (to be fed into the model as tensors) is represented by four letters:

\begin{itemize}
    \item N: Batch size, number of input processed together by the model
    \item C: Channel, $3$ for computer vision models
    \item W: Width, number of pixels in horizontal dimension
    \item H: Height, number of pixels in vertical dimension
\end{itemize}
Models are trained with input in either \texttt{NCHW} or \texttt{NHWC} data layout.
Figure~\ref{fig:platapus} shows the Top1 inference results of TensorFlow Inception v3 using different layouts for the same input image. The model was trained with \texttt{NHWC} layout. As can be seen, the predictions are very different.

\subsubsection{Image Decoding}

It is typical for authors to use JPEG as the image data format (with ImageNet being stored as JPEG images). There are different decoding methods for JPEG. One usually use \texttt{opencv.imread} or \texttt{PIL.Image.open} or \texttt{tf.image.decode\_jpeg} to decode a jpeg image. TensorFlow uses libJPEG and uses either \texttt{INTEGER\_FAST} or \texttt{INTEGER\_ACCURATE} as default (varies across systems); PIL maps to \texttt{INTEGER\_ACCURATE} method while OpenCV may not use libJPEG.

Even for the same method, ML libraries may have different implementations. For example, JPEG is stored on disk in YCrCb format, and the standard does not require bit-by-bit decoding accuracy. The implementation is defined differently across libraries, as shown in Figure~\ref{fig:pil_vs_opencv_impl}. Figure~\ref{fig:pil_vs_opencv_diff} shows the difference between decoding an image using Python Imaging Library (PIL) and OpenCV. We find that edge pixels (having high or low intensity) are not encoded consistently across libraries, even though these are the more interesting pixels for vision algorithms such as object detection.

\begin{figure}
\centering
\includegraphics[width=0.5\textwidth]{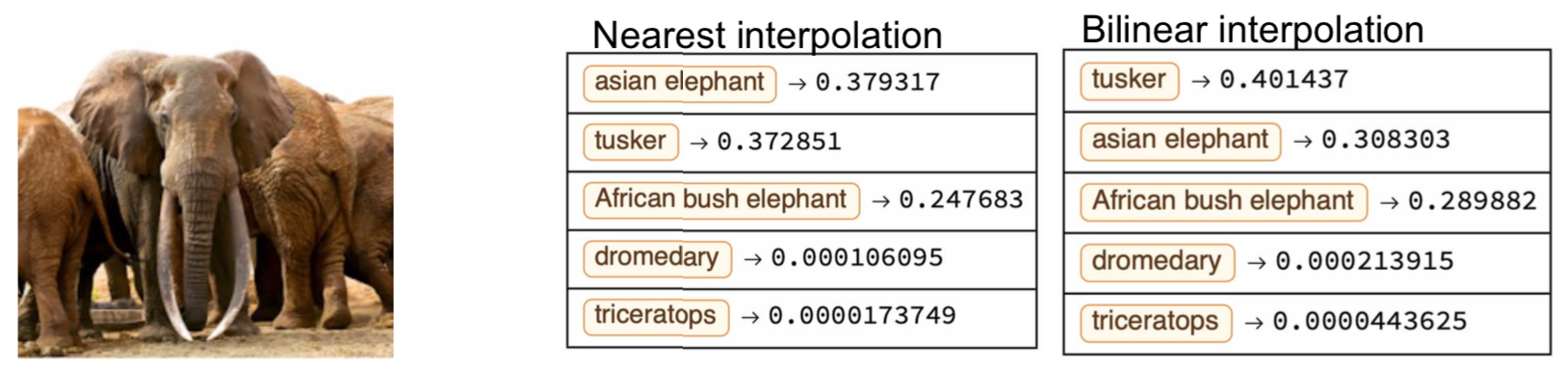}
\caption{Top 5 predictions using Inception-v3 with nearest interpolation or bilinear interpolation for resizing}
\label{fig:resizing_method}
\end{figure}

\begin{figure}
\centering
\includegraphics[width=0.5\textwidth]{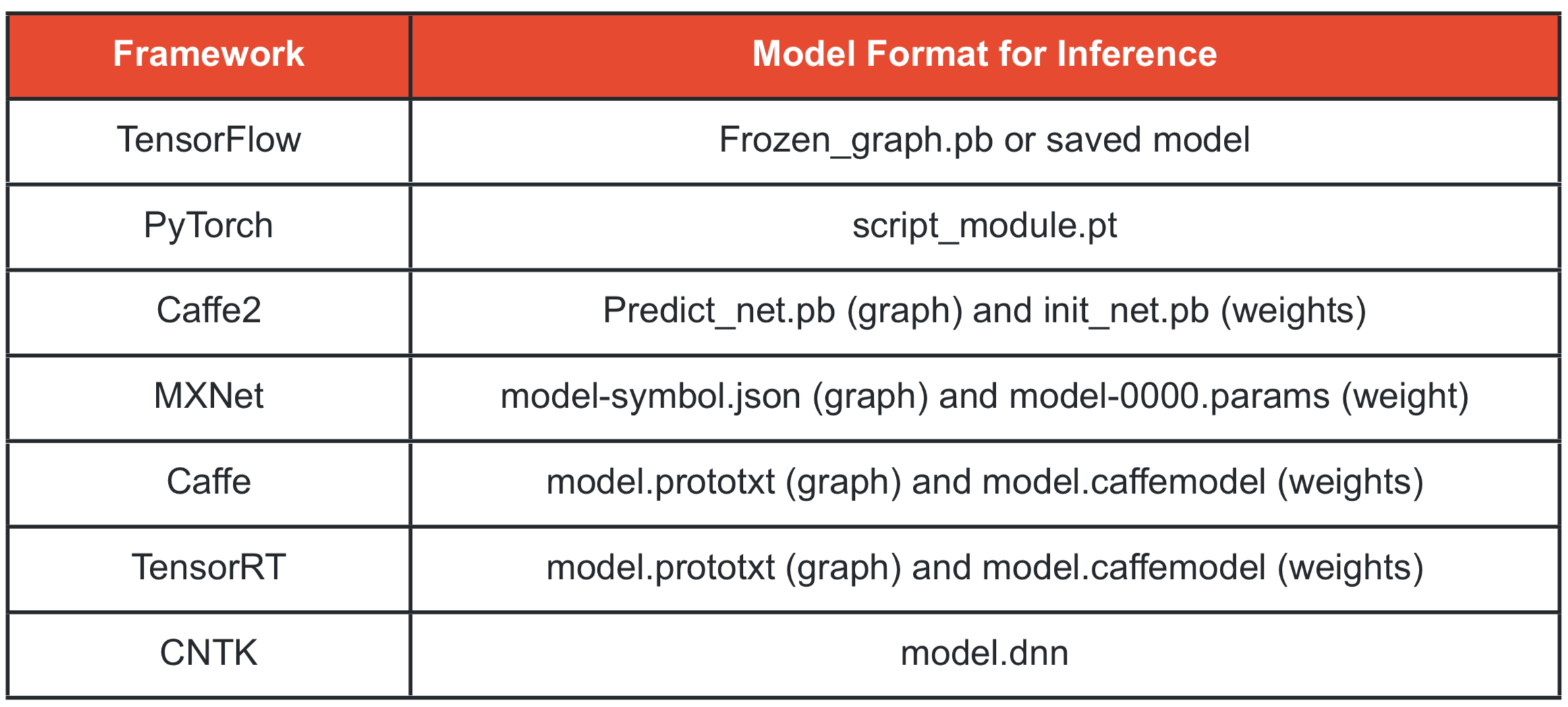}
\caption{Table of model inference formats for different frameworks.}
\label{fig:data_format_table}
\end{figure}

\begin{figure}
\centering
\includegraphics[width=0.5\textwidth]{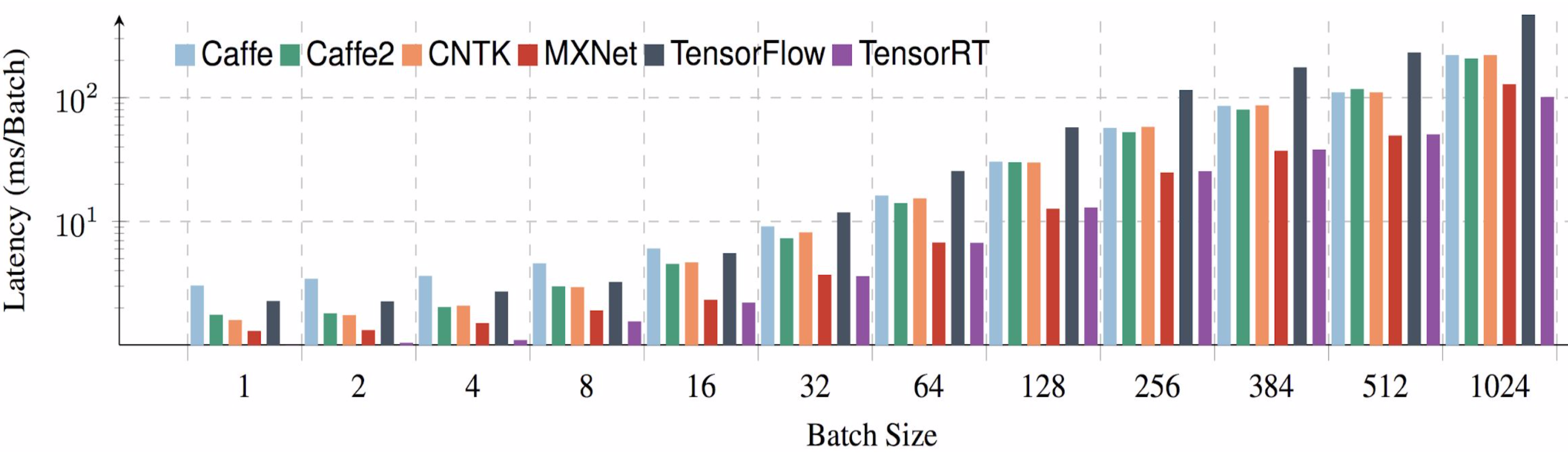}
\caption{AlexNet performance difference across frameworks on Volta.}
\label{fig:alexnet_machine_compare}
\end{figure}

\begin{figure*}[h]
\centering
\includegraphics[width=0.9\textwidth]{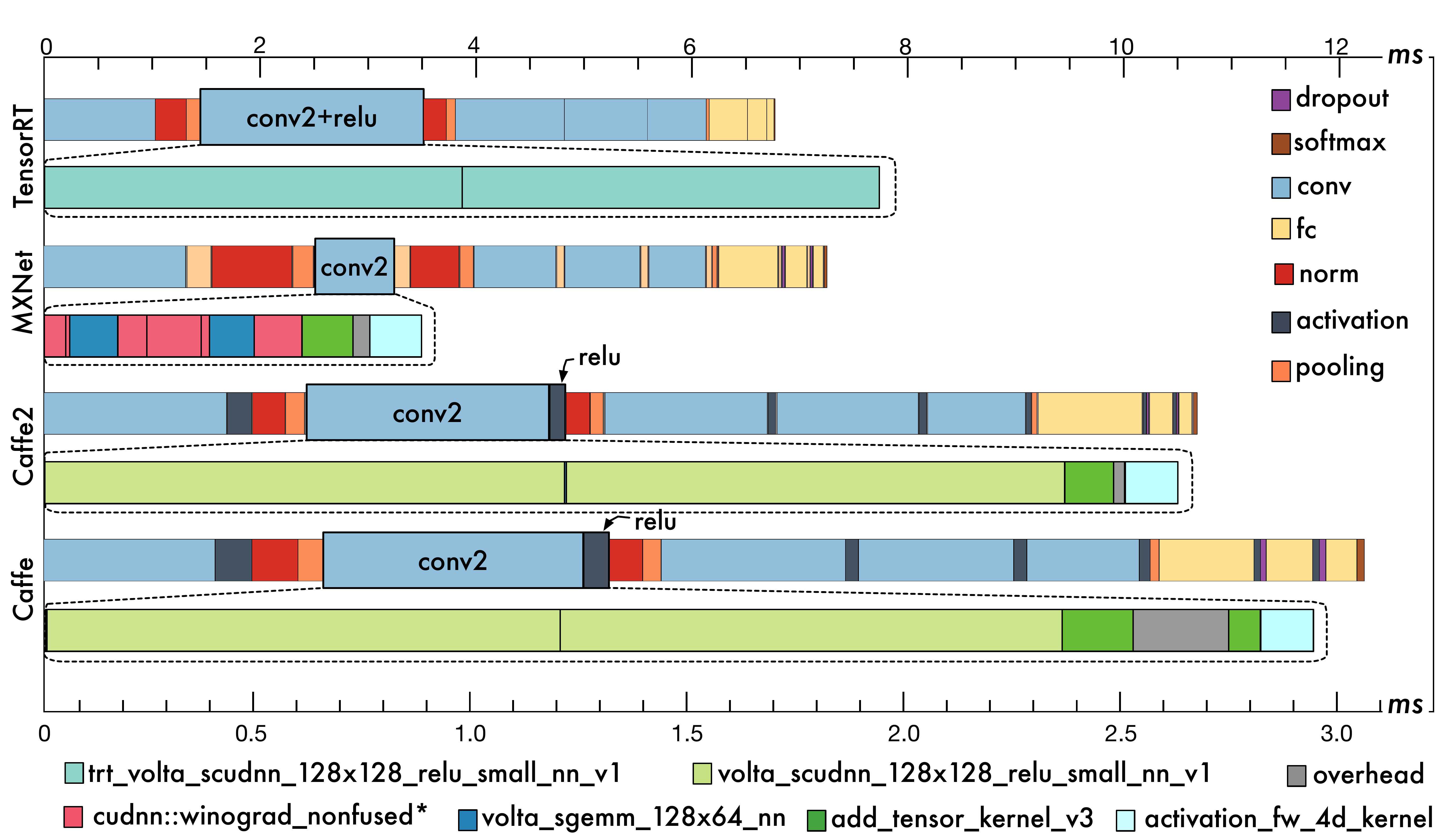}
\caption{Digging Deep into AlexNet Performance}
\label{fig:alexnet_framework_compare}
\end{figure*}

\subsubsection{Cropping}

For image classification, accuracy is sometimes reported for cropped validation datasets.
The cropping method and parameter are often overlooked by model evaluators, which results in different accuracy numbers. 
For Inception-v3, for example, the input images are center-cropped with fraction $87.5\%$, and then resized to $299\times299$. 
Figure~\ref{fig:diff_cropping} shows the effect of omitting cropping from pre-processing: (a) is the original image; (b) is the result of center cropping the image with $87.5\%$ and then resizing; (c) is the result of just resizing; (d) and (f) shows the prediction results using processed images from (b) and (c).
Intuitively, cropping differences are more pronounced for input images  where the marginal regions are meaningful (e.g. paintings within frames).

\subsubsection{Resizing Method}

Image input size is fixed per model, but resizing method is not widely described by model authors.
Multiple interpolation methods are available (nearest, bilinear, lanczos) and implementation for the same method can be different across libraries or frameworks.
Figure \ref{fig:resizing_method} shows the TensorFlow Inception-v3 Top 5 predictions difference between using nearest interpolation and using bilinear interpolation for resizing.

\subsubsection{Type Conversion and Normalization}

After decoding, the image data is in bytes and is converted to FP32 (assuming FP32 model) before being fed to the model. Also we need to subtract mean and scale the image data so that it has zero mean and unit variance ($(pixel-mean)/adjustedstddev$). 
Mathematically, float to byte conversion is float to byte conversion is $float\_to\_byte(x) = 255x$, and byte to float conversion is $byte\_to\_float(x) = x/255.0$.
Because of programming language semantics the executed behavior of byte to float conversion is $byte\_to\_float(x) = \floor{255x}$.

As part of the pre-processing, the input may also need to be normalized to have zero mean and unit variance ($\frac{pixel-mean}{stddev}$). 
We find that the order of operations for type conversion and normalization matters.
Figure~\ref{fig:diff_output_process} shows the image processing results using different order of operations for $meanByte = stddevByte = 127.5$ and $meanFloat = stddevFloat = 0.5$ where: (a) is the original image, (b) is the result of reading the image in bytes then normalizing it with the mean and standard deviation in bytes, $byte2float(\frac{imgByte - meanByte}{stddevByte})$, (c) is the result of reading an image in floats then normalizing it with the mean and standard deviation in floats, $\frac{byte2float(imgByte) - meanFloat}{stddevFloat}$, and (d) is the difference between (b) and (c)\footnote{\small To increase the contrast of the differences on paper, we dilate the image (with radius $2$) and rescaled its pixel values to cover the range between $0$ and $1$.}. 
The inference results of Figure~\ref{fig:diff_output_process} (b,c) are shown in Figure~\ref{fig:diff_output_process} (e,f).\abdul{this is not using tensorflow, but mxnet}

\subsection{Model and Data Formats}

There are a variety of formats used by ML frameworks to store models and data on disk, for inference and training. 
Some frameworks define models as Protocol Buffer~\cite{protobuf} and other use custom data formats. Figure~\ref{fig:data_format_table} shows the model format used for inference for different frameworks.
Some data formats such as TensorFlow TFRecord~\cite{tfrecord} or MXNet’s RecordIO~\cite{recordio} are optimized for static datasets. One can achieve 7x speedup with TFRecord and TF Dataset Iterator API for ImageNet evaluation.

\pitfall{Pitfall 4: Use an inappropriate format when measuring end-to-end performance.}

\subsection{Software Stack}

The major software components affecting reproducibility are ML framework (TensorFlow, MXNet, PyTorch, etc.) and libraries (MKL-DNN, Open-BLAS, cuDNN, etc.). They both impact not only the performance but also the accuracy of the model.

\pitfall{Pitfall 5a: If framework A and B use the same cuDNN and other ML libraries, they give the same performance and accuracy for the same model.}

Figure~\ref{fig:alexnet_machine_compare} shows AlexNet performance across different frameworks. All the frameworks are compiled with GCC 5.5 and use the same software stack (cuDNN and other libraries), but the performance is very different. With \carml, we can dig deeper into the inference processes of the frameworks to identify the bottlenecks and overheads of each framework.
Figure~\ref{fig:alexnet_framework_compare} shows that ML layers across frameworks have different implementations or dispatch to different library functions. Take the \texttt{conv2} and the following \texttt{relu} layers for example. In TensorRT, these two layers are merged together and are mapped to 2 \texttt{trt\_volta\_scudnn\_128x128\_relu\_small\_nn\_v1} kernels. While in other three frameworks, the two layers are not merged. Also the \texttt{conv2} layer in MXNet is executed very differently from the other frameworks.

\pitfall{Pitfall 5b: Same version of framework gives same performance.}

Framework installation and compilation affect model performance.
Benchmark results should report numbers with frameworks installed from source (with optimal compilation flags) for fair comparison.
Researchers usually have the choice to install a ML framework from source or from binary. Even through installation from binary is much easier, binary versions of framework may not use the CPU vectorization instructions (e.g. AVX, AVX2).
For example, TensorFlow 1.13 with vectorization is $40\%$ faster than one without for Inception-v3 using batch size 1 on CPUs.

\begin{figure}
\centering
\includegraphics[width=0.5\textwidth]{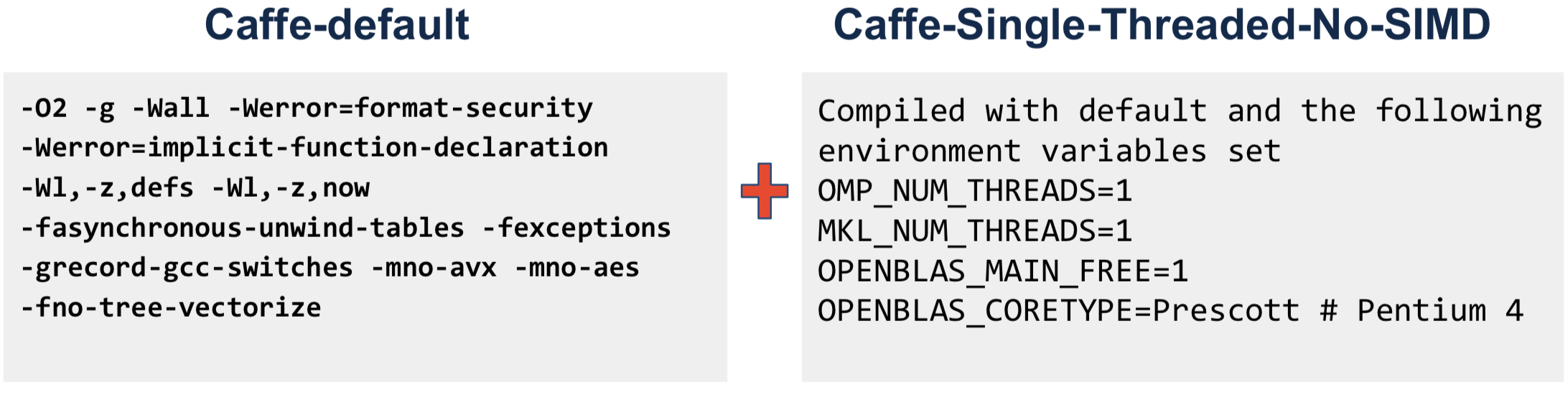}
\caption{Caffe installation options}
\label{fig:caffe_install_opts}
\end{figure}

\begin{figure}
\centering
\includegraphics[width=0.5\textwidth]{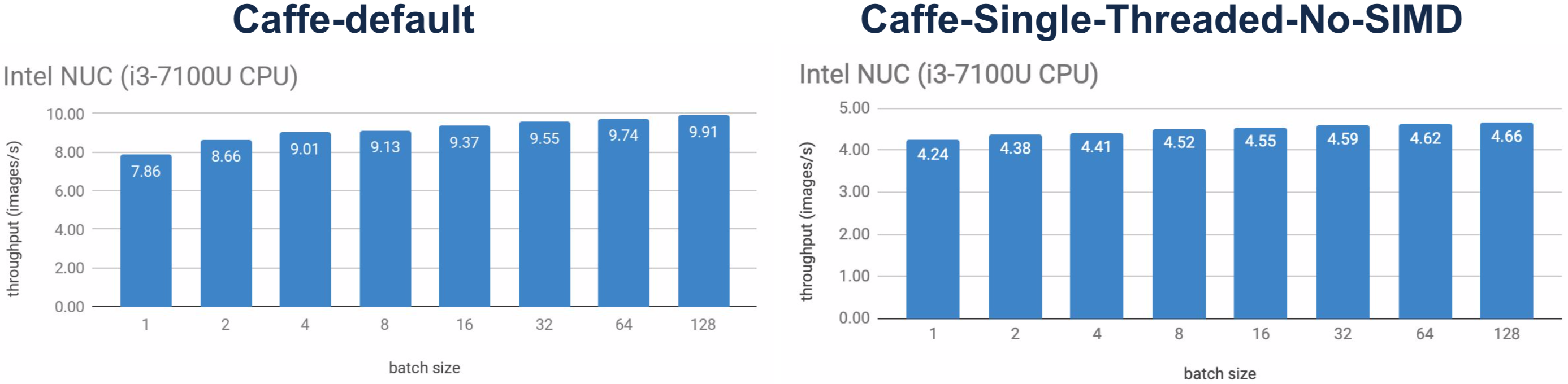}
\caption{Performance comparison between Caffe compiled with different options.}
\label{fig:caffe_install_diff_plot}
\end{figure}

Compilation options for framework and underlying libraries matters. For example, we compile Caffe using GCC 5.5 and with (1) the Caffe-default compiler flags in Figure~\ref{fig:caffe_install_opts}; (2) the Caffe-default and the Caffe-Single-Threaded-No-SIMD environment variables in Figure~\ref{fig:caffe_install_opts}.
Figure~\ref{fig:caffe_install_diff_plot} shows the SphereFace-20 performance comparison on a Intel NUC system between the two Caffe installations. As can be seen, Caffe-default is almost $2\times$ more performant than the other due to multithreading and vectorization.

\begin{figure}
\centering
\includegraphics[width=0.5\textwidth]{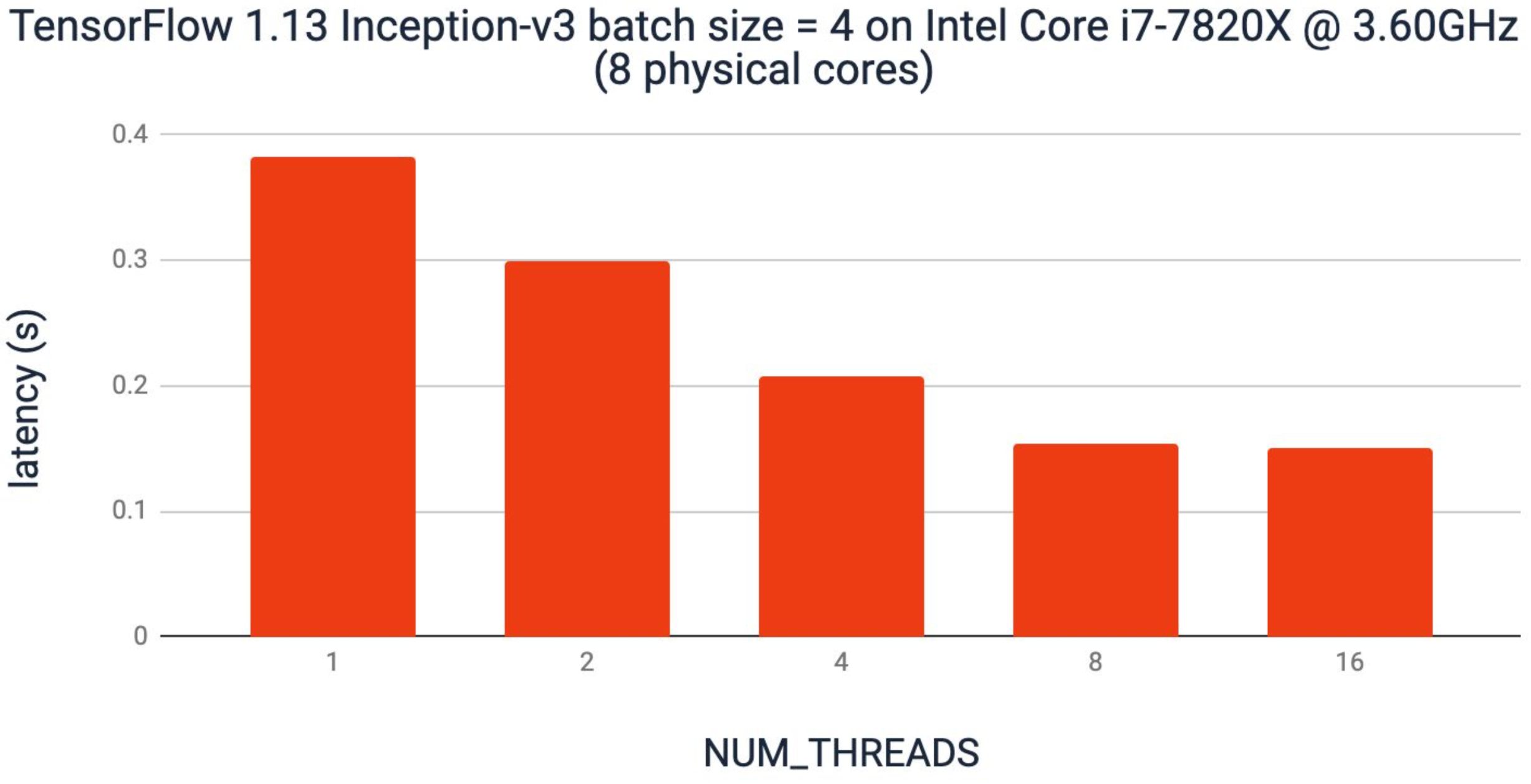}
\caption{TensorFlow inference latency varying num threads}
\label{fig:tensorflow_vary_num_threads}
\end{figure}

\subsection{Hardware Configuration}

Hardware configurations such as CPU Scaling, Multi-threading, Vectorization, affect mode evaluation performance. 

\pitfall{Pitfall 6: Always use the default hardware configurations without tuning the system for performance.}

Take Multi-threading for example
Modern CPUs have simultaneous multi-threading (also known as SMT or Hyper-threading). This allows multiple threads to run on the same core with the idea that each thread will not fully utilize the ALUs. As a study we vary the number of threads run by the TensorFlow using experiment variables \texttt{intra\_op\_parallelism\_threads} and \texttt{inter\_op\_parallelism\_threads}. The defaults for the two variables are the number of logical CPU cores and are effective for systems ranging from CPUs with $4$ to $70+$ combined logical cores~\cite{tfperf}. Figure~\ref{fig:tensorflow_vary_num_threads} shows Inception-v3 performance using different number of threads. On the system used which has 16 logical cores and 2-way SMT, the performance varies with different number of threads and the best is achieved using 16.

\section{Conclusion}

This article discusses some of the common pitfalls that one can encounter when trying to reproduce model evaluation.
To address these outlined challenges and pitfalls, we propose \carml in ~\cite{dakkak2018mlmodelscope}, an 
effective system solution for specifying and running model evaluation. \carml addresses the challenges and pitfalls with a model specification (referred to as model manifest), a distributed runtime to set up the required environments for running model evaluations, and an aggregation and summarization pipeline that captures application traces and explains the model execution process. To learn more, visit \hyperlink{http://mlmodelscope.org}{mlmodelscope.org}.

\bibliographystyle{plain}
\bibliography{references}
\end{document}